\documentclass[letterpaper, 10pt, conference]{ieeeconf}

\IEEEoverridecommandlockouts%

\makeatletter

\let\NAT@parse\relax
\makeatother

\usepackage[utf8]{inputenc}
\usepackage[T1]{fontenc}

\usepackage[english]{babel}

\usepackage{microtype}

\usepackage{
  amsfonts,
  amsmath,
  amssymb,
  amsthm,
  bm,
  mathtools
}

\usepackage{
  etoolbox,
  mathtools
}

\mathtoolsset{showonlyrefs,showmanualtags}

\mathtoolsset{mathic}

\providecommand\given{}

\DeclarePairedDelimiterX\set[1]{\{}{\}}{%
  \renewcommand\given{\SetSymbol[\delimsize]}%
  \ifblank{#1}{\:\cdots\:}{#1}
}

\DeclarePairedDelimiterX{\abs}[1]{\vert}{\vert}{
  \ifblank{#1}{\:\cdot\:}{#1}
}

\DeclarePairedDelimiterX{\norm}[1]{\lVert}{\rVert}{
  \ifblank{#1}{\:\cdot\:}{#1}
}
\DeclarePairedDelimiterXPP{\lnorm}[1]{}{\lVert}{\rVert}{_{2}}{
  \ifblank{#1}{\:\cdot\:}{#1}
}
\DeclarePairedDelimiterXPP{\infnorm}[1]{}{\lVert}{\rVert}{_{\infty}}{
  \ifblank{#1}{\:\cdot\:}{#1}
}
\DeclarePairedDelimiterXPP{\pnorm}[2]{}{\lVert}{\rVert}{
  \ifblank{#2}{_{p}}{_{#2}}}{
  \ifblank{#1}{\:\cdot\:}{#1}
}

\DeclarePairedDelimiterX{\inner}[2]{\langle}{\rangle}{
  \ifblank{#1}{\:\cdot\:}{#1},\ifblank{#2}{\:\cdot\:}{#2}
}

\DeclarePairedDelimiterXPP{\prob}[1]{\mathbb{P}}{[}{]}{}{
  \renewcommand\given{\nonscript\:\delimsize\vert\nonscript\:\mathopen{}}
  \ifblank{#1}{\:\cdot\:}{#1}
}

\DeclarePairedDelimiterXPP{\expv}[1]{\mathbb{E}}{[}{]}{}{
  \renewcommand\given{\nonscript\:\delimsize\vert\nonscript\:\mathopen{}}
  \ifblank{#1}{\:\cdot\:}{#1}
}

\usepackage{listings}

\usepackage[linesnumbered,vlined,ruled,commentsnumbered]{algorithm2e}
\SetVlineSkip{0.4em}
\DontPrintSemicolon
\SetKwIF{If}{ElseIf}{Else}{if}{}{else if}{else}{endif}
\SetAlgoSkip{}
\SetAlCapHSkip{0cm}

\usepackage{caption}
\usepackage{subcaption}
\usepackage{graphicx}
\usepackage{adjustbox}

\usepackage{tikz}
	\usetikzlibrary{calc,positioning}
	\tikzset{>=latex}
\usepackage{pgfplots, pgfplotstable}
\usepgfplotslibrary{statistics,fillbetween}
  \pgfplotsset{
    compat=1.14,
    mps basic/.style={
      xlabel near ticks,
      xlabel style={font=\footnotesize},
      ylabel near ticks,
      ylabel style={font=\footnotesize},
      xmajorgrids,
      major x grid style={dotted},
      ymajorgrids,
      major y grid style={dotted},
      tick label style={font=\footnotesize}
    },
    mps scientific x/.style={
      x tick label style={
        /pgf/number format/sci
      }
    },
    mps scientific y/.style={
      y tick label style={
        /pgf/number format/sci
      }
    },
    mps fixed x/.style={
      x tick label style={
        /pgf/number format/.cd,
        fixed,
        fixed zerofill,
        precision=6,
        /tikz/.cd
      }
    },
    mps fixed y/.style={
      y tick label style={
        /pgf/number format/.cd,
        fixed,
        fixed zerofill,
        precision=6,
        /tikz/.cd
      }
    }
  }

\usepackage{xcolor}
\definecolor{mpsgray}{RGB}{100,100,100}
\definecolor{mpsblue}{RGB}{11,93,174}
\definecolor{mpsred}{RGB}{206,62,21}
\definecolor{mpsyellow}{RGB}{232,163,26}
\definecolor{mpsgreen}{RGB}{100,161,27}
\definecolor{mpspurple}{RGB}{106,20,125}
\definecolor{mpslightblue}{RGB}{59,175,236}
\definecolor{mpsdarkred}{RGB}{145,0,33}
\usepackage{etoolbox}

\newcommand{\N}{\mathbb{N}}
\newcommand{\R}{\mathbb{R}}

\newcommand{\Rposinfty}{\mathbb{R}_{\geq 0}^{\infty}}

\newcommand{\SEthree}{\mathrm{SE}(3)}

\newcommand{\defeq}{\vcentcolon=}

\newcommand{\lebesgue}[1]{\lambda{\left(\ifblank{#1}{\:\cdot\:}{#1}\right)}}

\newcommand{\gtstandalone}{g_{\mathcal{T}}}
\newcommand{\gt}[1]{\gtstandalone\left(#1\right)}
\newcommand{\ghatstandalone}{\widehat{g}}
\newcommand{\ghat}[1]{\ghatstandalone\left(#1\right)}
\newcommand{\chatstandalone}{\widehat{c}}
\newcommand{\chat}[2]{\chatstandalone\left(#1, #2\right)}
\newcommand{\ctruestandalone}{c}
\newcommand{\ctrue}[2]{\ctruestandalone\left(#1, #2\right)}
\newcommand{\hhatstandalone}{\widehat{h}}
\newcommand{\hhat}[1]{\hhatstandalone\left(#1\right)}
\newcommand{\fhatstandalone}{\widehat{f}}
\newcommand{\fhat}[1]{\fhatstandalone\left(#1\right)}

\newcommand{\sspace}{X}
\newcommand{\ssampled}{X_{\mathrm{sampled}}}

\newcommand{\sinformed}{X_{\fhatstandalone}}

\newcommand{\sstart}{\bm{\mathrm{x}}_{\mathrm{init}}}
\newcommand{\sgoal}{\bm{\mathrm{x}}_{\mathrm{goal}}}
\newcommand{\sgoals}{X_{\mathrm{goal}}}
\newcommand{\sany}{\bm{\mathrm{x}}}
\newcommand{\snum}[1]{\bm{\mathrm{x}}_{#1}}
\newcommand{\sparent}{\bm{\mathrm{x}}_{\mathrm{p}}}
\newcommand{\schild}{\bm{\mathrm{x}}_{\mathrm{c}}}

\newcommand{\sset}{\leftarrow{}}
\newcommand{\ssetinsert}{\overset{+}{\leftarrow}}
\newcommand{\ssetremove}{\overset{-}{\leftarrow}}

\usepackage[numbers,sort&compress]{natbib}

\newcommand{\updateheuristic}[1]{\texttt{update\_heuristic}\left(#1\right)}
\newcommand{\mpsupdatestate}[1]{\texttt{update\_state}\left(#1\right)}
\newcommand{\popbest}[1]{\texttt{pop\_best}\left(#1\right)}
\newcommand{\isvalid}[1]{\texttt{is\_valid}\left(#1\right)}
\newcommand{\mpsexpand}[1]{\texttt{expand}\left(#1\right)}
\newcommand{\mpsfparent}[1]{\texttt{parent}_{\mathcal{F}}\left(#1\right)}
\newcommand{\mpsrparent}[1]{\texttt{parent}_{\mathcal{R}}\left(#1\right)}
\newcommand{\mpsfchildren}[1]{\texttt{children}_{\mathcal{F}}\left(#1\right)}
\newcommand{\mpsrchildren}[1]{\texttt{children}_{\mathcal{R}}\left(#1\right)}
\newcommand{\mpssample}[1]{\texttt{sample}\left(#1\right)}
\newcommand{\mpsneighbors}[1]{\texttt{neighbors}\left(#1\right)}
\renewcommand{\gtstandalone}{g_{\mathcal{F}}}
\renewcommand{\gt}[1]{\gtstandalone\left(#1\right)}
\newcommand{\hhatconstandalone}{\widehat{h}_{\mathrm{con}}}
\newcommand{\hhatcon}[1]{\hhatconstandalone\left[#1\right]}
\newcommand{\hhatexpstandalone}{\widehat{h}_{\mathrm{exp}}}
\newcommand{\hhatexp}[1]{\hhatexpstandalone\left[#1\right]}
\newcommand{\fqueue}{\mathcal{Q}_{\mathrm{F}}}
\newcommand{\rqueue}{\mathcal{Q}_{\mathrm{R}}}
\newcommand{\ftree}{\mathcal{F}}
\newcommand{\rtree}{\mathcal{R}}
\newcommand{\fvertices}{V_{\mathcal{F}}}
\newcommand{\rvertices}{V_{\mathcal{R}}}
\newcommand{\fedges}{E_{\mathcal{F}}}
\newcommand{\redges}{E_{\mathcal{R}}}
\newcommand{\edgeblacklist}{E_{\mathrm{invalid}}}
\newcommand{\aitstar}{AIT*}

\pgfplotsset{compat=1.14}

\usepackage{bold-extra}

\SetVlineSkip{0.2em}

\usetikzlibrary{external}
\usepackage[hidelinks]{hyperref}

\title{\LARGE \bf
  Adaptively Informed Trees (\aitstar):\\Fast Asymptotically Optimal Path Planning through Adaptive Heuristics
}

\author{Marlin P.\ Strub$^{1}$ and Jonathan D.\ Gammell$^{1}$%
  \thanks{$^{1}$M.\ P.\ Strub and J.\ D.\ Gammell are with the Estimation, Search, and Planning (ESP) Group of the Oxford Robotics Institute (ORI), University of Oxford, United Kingdom. {\tt\footnotesize (mstrub|gammell)@robots.ox.ac.uk}}%
}

\begin{document}

\maketitle
\thispagestyle{empty}
\pagestyle{empty}

\begin{abstract}

  Informed sampling-based planning algorithms exploit problem knowledge for better search performance. This knowledge is often expressed as heuristic estimates of solution cost and used to order the search. The practical improvement of this informed search depends on the accuracy of the heuristic.

  Selecting an appropriate heuristic is difficult. Heuristics applicable to an entire problem domain are often simple to define and inexpensive to evaluate but may not be beneficial for a specific problem instance. Heuristics specific to a problem instance are often difficult to define or expensive to evaluate but can make the search itself trivial.

  This paper presents Adaptively Informed Trees (\aitstar{}), an almost-surely asymptotically optimal sampling-based planner based on BIT*. AIT* adapts its search to each problem instance by using an asymmetric bidirectional search to simultaneously estimate and exploit a problem-specific heuristic. This allows it to quickly find initial solutions and converge towards the optimum. \aitstar{} solves the tested problems as fast as RRT-Connect while also converging towards the optimum.

\end{abstract}

\bstctlcite{IEEEexample:BSTcontrol} 

\section{Introduction}%
\label{sec:introduction}

Path planning is the problem of finding a continuous sequence of valid states between a start and goal specification. Sampling-based planners, such as Probabilistic Roadmaps (PRM)~\citep{kavraki1996}, approximate the state space by sampling discrete states and connecting them with edges. The resulting structure can then be processed by graph-search algorithms to find a sequence of states that connects the start to the goal.

Informed graph-search algorithms, such as A*~\citep{hart1968}, use knowledge about a problem domain to increase their efficiency. This knowledge is often captured in the form of a \emph{heuristic function}, \( \hhatstandalone \), which estimates cost-to-go, i.e., the cost to go from any state in the state space to the goal. 

The properties of this heuristic directly affect the performance of the search algorithms. An \emph{admissible} heuristic never overestimates the actual cost-to-go. A \emph{consistent} heuristic satisfies a triangle-inequality, such that for any two states, \( \snum{i}, \snum{j} \), it satisfies \( \hhat{\snum{i}} \leq p(\snum{i}, \snum{j}) + \hhat{\snum{j}} \), where \( p(\snum{i}, \snum{j}) \) is the best cost of any path from \( \snum{i} \) to \( \snum{j} \). Note that by definition all consistent heuristics are also admissible. A* is guaranteed to find the optimal solution when provided with an admissible heuristic. If the provided heuristic is also consistent, then A* expands the minimum number of states of any informed graph-search algorithm using that heuristic (i.e., it is \emph{optimally efficient}~\citep{hart1968}).

Improving the accuracy of a heuristic directly improves the performance of informed search algorithms~\citep{korf1997,culberson1998,felner2004,paden2017}, and the search becomes trivial when a perfect heuristic is available.

Designing and selecting effective heuristics is difficult for many problem domains. This is because heuristics are most effective when they are both accurate and computationally inexpensive to evaluate. Heuristics that are applicable to an entire problem domain are often inexpensive to evaluate but may not be accurate for a specific problem instance. Accurate heuristics can be designed for a specific problem instance during the search, but this can be computationally expensive and may diminish the overall search performance.

\begin{figure}[t]
  \centering
  \includegraphics[width = \columnwidth]{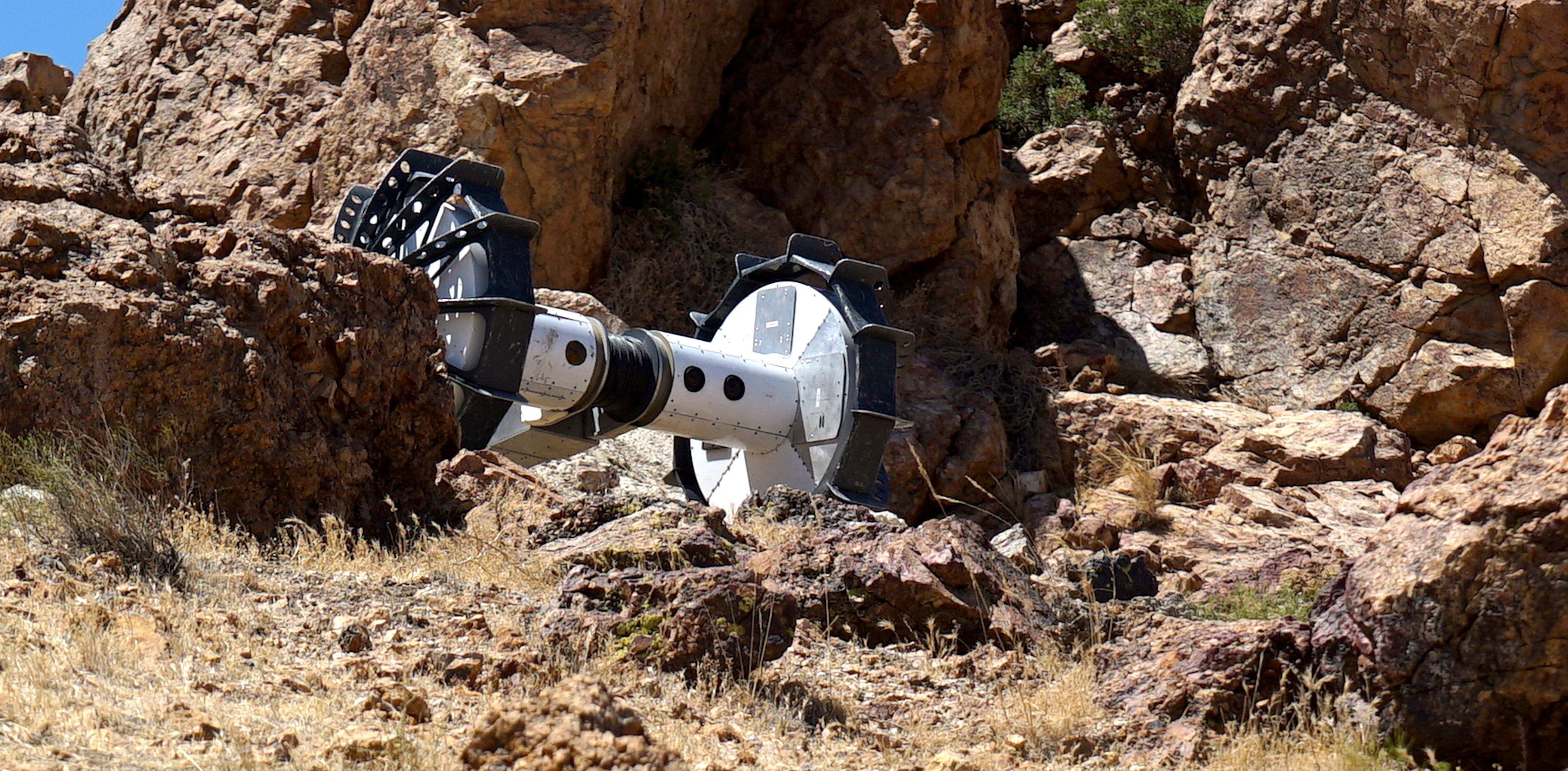}
  \caption{AIT* uses an asymmetric bidirectional search that is specialized for path planning problems with computationally expensive edge evaluations, such as those posed by NASA/JPL-Caltech's Axel Rover System~\citep{nesnas2012}.%
}%
\label{fig:teaser}
\end{figure}%

Computational cost directly influences the real-world performance of planning algorithms. Sampling-based planners contain a number of computationally expensive basic operations, including state expansions and edge evaluations~\citep{kleinbort2016}. State expansions often require nearest neighbor searches that increase in computational cost with the number of samples. Edge evaluations require local planning between two states and detecting collisions on the resulting path.

Lazy sampling-based planners, such as Lazy PRM~\citep{bohlin2000}, reduce this computational cost by avoiding the evaluation of every edge. These algorithms first perform an inexpensive search on a simplified approximation without collision detection. This allows them to only evaluate the edges that are believed to be on an optimal path, and reduce the number of evaluated edges. This improves performance, especially for problems with computationally expensive edge evaluations, such as those considered in this paper.

This paper presents Adaptively Informed Trees (\aitstar{}), a lazy, almost-surely asymptotically optimal sampling-based planner that uses an asymmetric bidirectional search to simultaneously estimate and exploit an accurate, problem-specific heuristic. \aitstar{} estimates this heuristic by performing a lazy reverse search on the current sampling-based approximation. This heuristic is then used to order the forward search of this approximation while considering complete edge evaluations. The results of the computationally expensive edge evaluations performed by this forward search inform the reverse search, which creates increasingly accurate heuristics. This allows \aitstar{} to efficiently share information between the two individual searches. 

Efficiently estimating and exploiting a problem-specific heuristic allows \aitstar{} to outperform existing sampling-based planning algorithms when edge evaluations are expensive. \aitstar{} finds initial solutions to the tested problems at least as fast as RRT-Connect while still almost-surely converging to the optimal solution, which RRT-Connect does not. 

\section{Background}%
\label{sec:background}

Creating and adapting heuristics to improve informed graph-search algorithms is an active area of research (Sec.~\ref{sec:improved-heuristics-for-informed-search-algorithms}). Heuristics are applied in sampling-based planning to reduce search effort by ordering the search and focus the approximation to the relevant region of the state space (Sec.~\ref{sec:sampling-based-path-planning-with-heuristics}). Lazy search algorithms separately focus on reducing search effort by avoiding collision detection~(Sec.~\ref{sec:sampling-based-path-planning-with-lazy-collision-detection}).

\subsection{Improved Heuristics for Informed Search Algorithms}%
\label{sec:improved-heuristics-for-informed-search-algorithms}

Improving the heuristic for informed graph-search algorithms has been shown to be effective for many problem domains, including path planning~\citep{paden2017}.

Pattern Databases~\citep{culberson1998} are precomputed tables of exact solution costs to simplified subproblems of a problem domain. An informed algorithm can use this database during the search to create admissible heuristics. Additive Pattern Databases~\citep{felner2004} extend this approach to combine database entries into more accurate heuristics that are still admissible. This approach speeds up the search of problems for which simplified subproblems can be created and solved, but requires creating databases for every problem domain \textit{a priori} to the search.

Heuristic accuracy can alternatively be improved by using the error in the heuristic values of states as they are discovered. Thayer et al.~\citep{thayer2011} use the error of each state expansion to update the heuristic during the search. This can be applied to any problem domain and does not require any preprocessing, but the resulting heuristic is not guaranteed to be admissible.

Adaptive A*~\citep{koenig2005,koenig2006a} is an incremental search algorithm that updates its heuristic function based on the cost-to-come values of previous searches of similar problems. This results in ever more accurate and admissible heuristics but can not be used for the initial search of a graph.

Kaindl. et al~\citep{kaindl1997} use the \emph{Add method} to inform a forward search with a partial reverse search. The reverse search reveals errors in the heuristic values, the minimum of which is added to all unexpanded states. This results in a more informed but still admissible heuristic, but requires the user to specify how many states to expand during the reverse search and increases the heuristic uniformly for all unexpanded states. Wilt et al.~\citep{wilt2013} present an updated version of this method which does not require a user-specified parameter.

Unlike these approaches, \aitstar{} does not need a predefined database, creates a consistent heuristic during the search, can be used on the initial search of a graph, and adaptively estimates the heuristic for each state individually.

\subsection{Sampling-Based Planning with Heuristics}%
\label{sec:sampling-based-path-planning-with-heuristics}

Heuristics have been used in sampling-based planning to guide the search and focus the approximation.  RRT-Connect~\citep{kuffner2000} builds on Rapidly-exploring Random Trees (RRT)~\citep{lavalle2001} by incrementally growing two trees, one rooted in the start state and one in the goal state. These trees each explore the state space around them but are also guided towards each other by a \emph{connect heuristic}. This approach can result in very fast initial solution times but is not almost-surely asymptotically optimal and does not improve the solution quality with more computational time.

Informed RRT*~\citep{gammell2018} incorporates an ellipsoidal heuristic into the almost-surely asymptotically optimal RRT*~\citep{karaman2011}. This improves the convergence rate by focusing the incremental approximation to the relevant region of the state space but does not guide the search.

Sakcak et al.~\citep{sakcak2019a} incorporate a heuristic into a version of RRT* which is based on motion-primitives~\citep{sakcak2019b}. This can improve the performance for kinodynamic systems but requires preprocessing and relies on an \textit{a priori} discretization which suffers from the \textit{curse of dimensionality}~\citep{bellman1957}. 

Batch Informed Trees (BIT*)~\citep{gammell2015,gammell2020} samples batches of states and views these sampled states as an increasingly dense edge-implicit random geometric graph (RGG)~\citep{penrose2003}. This allows BIT* to use a series of informed graph-searches to process the states in order of potential solution quality. BIT* efficiently reuses information from both previous searches and approximations by using incremental search techniques but does not update its heuristic during the search.

Unlike these approaches, \aitstar{} improves its solution quality with more computational time, uses its heuristic to guide the search, does not rely on an \textit{a priori} discretization, and updates its heuristic during the search.

\subsection{Sampling-Based Planning with Lazy Collision Detection}%
\label{sec:sampling-based-path-planning-with-lazy-collision-detection}

Path planning algorithms employ lazy collision detection to avoid spending computational resources on edges that are unlikely to be on an optimal path. Lazy PRM~\citep{bohlin2000} approximates the entire state space with an RGG without collision detection and searches this RGG for a path from the start state to a goal state. This path is then checked for collisions. If collisions are detected, then the corresponding edges and vertices are removed from the graph and a new search must be started from scratch. There also exist almost-surely asymptotically optimal variants of Lazy PRM~\citep{hauser2015,kim2018}.

Lazy Shortest Path (LazySP)~\citep{dellin2016} is a class of algorithms that reduces the number of edges checked for collisions. It first finds a path from the start to the goal using an inexpensive estimate of the edge costs. Once a path is found, it uses an \textit{edge selector} function which determines the order in which these edges are checked for collision. An example of a LazySP algorithm is Lazy Receding Horizon A* (LRHA*)~\citep{mandalika2018}.

Unlike these approaches, \aitstar{} does not restart its search from scratch upon detecting collisions, and uses admissible heuristics to focus its approximation.

\section{Adaptively Informed Trees (AIT*)}%
\label{sec:adaptively-informed-trees}

BIT* is an almost-surely asymptotically optimal sampling-based planner that builds a discrete approximation of a state space by sampling batches of states. This approximation can be focused to the region of the state space that can possibly improve a current solution with informed sampling~\citep{gammell2018}.

BIT* views the states it samples as an increasingly dense, edge-implicit RGG.\@ It processes the implicit RGG edges in order of potential solution quality, similar to an edge-queue version of Lifelong Planning A* (LPA*)~\citep{koenig2004}. The true edge costs are evaluated lazily by maintaining a queue ordered by the sum of the current cost-to-come from the start to the edge's parent state, a heuristic of the edge cost, and a heuristic of the cost-to-go from the edge's child state to the goal state. Full details of BIT* are in~\citep{gammell2017,gammell2020}.

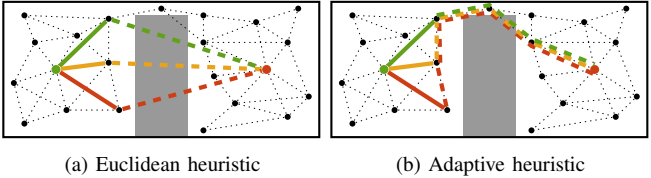
\begin{figure}[t]
  \begin{minipage}[b]{0.495\columnwidth}
      \begin{tikzpicture}[
    xscale=1.4, yscale=0.9,
    state/.style={circle, inner sep = 0.8pt, fill = black},
    edge/.style={black, very thin, dash pattern = on 0.5pt off 1.5pt}
  ]
  \draw [draw=none, fill=black!40] (-0.25, -1) rectangle (0.25, 0.8);
  \draw [black, thick] (-1.5, -1) rectangle (1.5, 1);
  \node (start) [state, inner sep = 1.2pt, fill=mpsgreen] at (-1.0, 0.0) {};
  \node (goal) [state, inner sep = 1.2pt, fill=mpsred] at (1.0, 0.0) {};
  \node (x1) [state] at (-0.5, 0.75) {};
  \node (x2) [state] at (-0.4, -0.6) {};
  \node (x7) [state] at (-0.5, 0.1) {};
  \node (x3) [state] at (-0.9, -0.6) {};
  \node (x4) [state] at (-1.4, -0.2) {};
  \node (x5) [state] at (-1.3, -0.8) {};
  \node (x8) [state] at (-0.8, 0.5) {};
  \node (x9) [state] at (-1.2, 0.4) {};
  \node (x11) [state] at (-1.3, 0.8) {};
  \node (x12) [state] at (0.0, 0.9) {};
  \node (x13) [state] at (0.4, 0.8) {};
  \node (x14) [state] at (1.3, 0.9) {};
  \node (x15) [state] at (0.9, 0.7) {};
  \node (x16) [state] at (0.4, 0.4) {};
  \node (x17) [state] at (1.2, 0.4) {};
  \node (x18) [state] at (0.5, -0.1) {};
  \node (x19) [state] at (0.7, -0.5) {};
  \node (x20) [state] at (1.4, -0.4) {};
  \node (x21) [state] at (1.2, -0.7) {};
  \node (x22) [state] at (0.4, -0.9) {};
  \draw [edge] (start) -- (x3);
  \draw [edge] (start) -- (x4);
  \draw [edge] (start) -- (x5);
  \draw [edge] (start) -- (x8);
  \draw [edge] (start) -- (x9);
  \draw [edge] (x2) -- (x7);
  \draw [edge] (x3) -- (x7);
  \draw [edge] (x4) -- (x5);
  \draw [edge] (x4) -- (x3);
  \draw [edge] (x4) -- (x9);
  \draw [edge] (x5) -- (x3);
  \draw [edge] (x3) -- (x2);
  \draw [edge] (x7) -- (x8);
  \draw [edge] (x7) -- (x1);
  \draw [edge] (x8) -- (x9);
  \draw [edge] (x8) -- (x11);
  \draw [edge] (x9) -- (x11);
  \draw [edge] (x1) -- (x11);
  \draw [edge] (x1) -- (x8);
  \draw [edge] (x1) -- (x12);
  \draw [edge] (x12) -- (x13);
  \draw [edge] (x12) -- (x16);
  \draw [edge] (x13) -- (x15);
  \draw [edge] (x13) -- (x16);
  \draw [edge] (x14) -- (x15);
  \draw [edge] (x14) -- (x17);
  \draw [edge] (x15) -- (x16);
  \draw [edge] (x15) -- (x17);
  \draw [edge] (x15) -- (goal);
  \draw [edge] (x16) -- (x17);
  \draw [edge] (x16) -- (x18);
  \draw [edge] (x16) -- (goal);
  \draw [edge] (x17) -- (x20);
  \draw [edge] (x17) -- (goal);
  \draw [edge] (x18) -- (goal);
  \draw [edge] (x18) -- (x19);
  \draw [edge] (x19) -- (goal);
  \draw [edge] (x19) -- (x20);
  \draw [edge] (x19) -- (x21);
  \draw [edge] (x19) -- (x22);
  \draw [edge] (x20) -- (goal);
  \draw [edge] (x20) -- (x21);
  \draw [edge] (x21) -- (x22);
  \draw [edge] (x21) -- (goal);
  \draw [edge, draw=mpsgreen, ultra thick, solid] (start) -- (x1);
  \draw [edge, draw=mpsyellow, ultra thick, solid] (start) -- (x7);
  \draw [edge, draw=mpsred, ultra thick, solid] (start) -- (x2);
  \draw [mpsgreen, dashed, ultra thick] (x1) -- (goal);
  \draw [mpsyellow, dashed, ultra thick] (x7) -- (goal);
  \draw [mpsred, dashed, ultra thick] (x2) -- (goal);
\end{tikzpicture}%
    \subcaption{Euclidean heuristic}%
    \label{fig:euclidean-heuristic}%
  \end{minipage}%
  \hfill%
  \begin{minipage}[b]{0.495\columnwidth}
      \begin{tikzpicture}[
    xscale=1.4, yscale=0.9,
    state/.style={circle, inner sep = 0.8pt, fill = black},
    edge/.style={black, very thin, dash pattern = on 0.5pt off 1.5pt}
  ]
  \draw [draw=none, fill=black!40] (-0.25, -1) rectangle (0.25, 0.8);
  \draw [black, thick] (-1.5, -1) rectangle (1.5, 1);
  \node (start) [state, fill=mpsgreen, inner sep = 1.2pt] at (-1.0, 0.0) {};
  \node (goal) [state, fill=mpsred, inner sep = 1.2pt] at (1.0, 0.0) {};
  \node (x1) [state] at (-0.5, 0.75) {};
  \node (x2) [state] at (-0.4, -0.6) {};
  \node (x7) [state] at (-0.5, 0.1) {};
  \node (x3) [state] at (-0.9, -0.6) {};
  \node (x4) [state] at (-1.4, -0.2) {};
  \node (x5) [state] at (-1.3, -0.8) {};
  \node (x8) [state] at (-0.8, 0.5) {};
  \node (x9) [state] at (-1.2, 0.4) {};
  \node (x11) [state] at (-1.3, 0.8) {};
  \node (x12) [state] at (0.0, 0.9) {};
  \node (x13) [state] at (0.4, 0.8) {};
  \node (x14) [state] at (1.3, 0.9) {};
  \node (x15) [state] at (0.9, 0.7) {};
  \node (x16) [state] at (0.4, 0.4) {};
  \node (x17) [state] at (1.2, 0.4) {};
  \node (x18) [state] at (0.5, -0.1) {};
  \node (x19) [state] at (0.7, -0.5) {};
  \node (x20) [state] at (1.4, -0.4) {};
  \node (x21) [state] at (1.2, -0.7) {};
  \node (x22) [state] at (0.4, -0.9) {};
  \draw [edge] (start) -- (x3);
  \draw [edge] (start) -- (x4);
  \draw [edge] (start) -- (x5);
  \draw [edge] (start) -- (x8);
  \draw [edge] (start) -- (x9);
  \draw [edge] (x2) -- (x7);
  \draw [edge] (x3) -- (x7);
  \draw [edge] (x4) -- (x5);
  \draw [edge] (x4) -- (x3);
  \draw [edge] (x4) -- (x9);
  \draw [edge] (x5) -- (x3);
  \draw [edge] (x3) -- (x2);
  \draw [edge] (x7) -- (x8);
  \draw [edge] (x7) -- (x1);
  \draw [edge] (x8) -- (x9);
  \draw [edge] (x8) -- (x11);
  \draw [edge] (x9) -- (x11);
  \draw [edge] (x1) -- (x11);
  \draw [edge] (x1) -- (x8);
  \draw [edge] (x1) -- (x12);
  \draw [edge] (x12) -- (x13);
  \draw [edge] (x12) -- (x16);
  \draw [edge] (x13) -- (x15);
  \draw [edge] (x13) -- (x16);
  \draw [edge] (x14) -- (x15);
  \draw [edge] (x14) -- (x17);
  \draw [edge] (x15) -- (x16);
  \draw [edge] (x15) -- (x17);
  \draw [edge] (x15) -- (goal);
  \draw [edge] (x16) -- (x17);
  \draw [edge] (x16) -- (x18);
  \draw [edge] (x16) -- (goal);
  \draw [edge] (x17) -- (x20);
  \draw [edge] (x17) -- (goal);
  \draw [edge] (x18) -- (goal);
  \draw [edge] (x18) -- (x19);
  \draw [edge] (x19) -- (goal);
  \draw [edge] (x19) -- (x20);
  \draw [edge] (x19) -- (x21);
  \draw [edge] (x19) -- (x22);
  \draw [edge] (x20) -- (goal);
  \draw [edge] (x20) -- (x21);
  \draw [edge] (x21) -- (x22);
  \draw [edge] (x21) -- (goal);
  \draw [edge, draw=mpsgreen, ultra thick, solid] (start) -- (x1);
  \draw [edge, draw=mpsyellow, ultra thick, solid] (start) -- (x7);
  \draw [edge, draw=mpsred, ultra thick, solid] (start) -- (x2);
  \draw [mpsred, dashed, ultra thick] (x2) -- (x7) (x7.east) -- (x1.south east) -- (x12.south) -- (x16.south) -- (goal.south);
  \draw [mpsyellow, dashed, ultra thick] (x7) -- (x1) -- (x12) -- (x16) -- (goal);
  \draw [mpsgreen, dashed, ultra thick] (x1.north) -- (x12.north) -- (x16.north) -- (goal.north);
\end{tikzpicture}%
    \subcaption{Adaptive heuristic}%
    \label{fig:adaptive-heuristic}%
  \end{minipage}
  \caption{BIT* approximates the state space of a problem by sampling batches of states (black dots), which define an RGG with implicit edges (dotted connections). These edges are processed in order of potential solution quality, according to a heuristic cost-to-go estimate (thick dashed lines). \textit{A priori} heuristics are often not problem specific and provide poor estimates of the cost-to-go. The Euclidean norm assigns the lowest potential solution cost to the yellow edge despite the presence of the obstacle~(\subref{fig:euclidean-heuristic}). By using a lazy reverse search, AIT* is instead able to provide a more accurate heuristic adapted to the current approximation of the specific problem~(\subref{fig:adaptive-heuristic}).}%
  \label{fig:heuristic-visualization}
\end{figure}

\aitstar{} builds on BIT* by improving the accuracy of the used heuristic, which improves performance on problems with expensive edge evaluations. It uses an asymmetric bidirectional search to efficiently estimate and exploit an accurate heuristic for each problem instance (Fig.~\ref{fig:heuristic-visualization}). The forward search is the same as in BIT* but uses the heuristic provided by a computationally inexpensive reverse search. This heuristic can be updated efficiently when the forward search reveals that it contains errors by using an incremental algorithm, such as LPA*, on the reverse search. Algorithm~\ref{alg:conceptual} presents a conceptual overview of BIT* and \aitstar{}. The full algorithmic details are provided in Algorithms~\ref{alg:technical}--\ref{alg:updatestate}.

Since BIT* is almost-surely asymptotically optimal when given an admissible heuristic~\citep{gammell2020} and the reverse search of \aitstar{} results in an admissible heuristic for each approximation, \aitstar{} is almost-surely asymptotically optimal as well.

\subsection{Notation}%
\label{sec:notation}

The state space of the planning problem is denoted by \( \sspace = {\R}^{n}, n \in \N \), the start by \( \sstart \in \sspace \), and the goals by \( \sgoals \in \sspace \). The sampled states are denoted by \( \ssampled \).

The forward and reverse search trees are denoted by \( \ftree = (\fvertices, \fedges) \), and \( \rtree = (\rvertices, \redges) \), respectively. The vertices in these trees, denoted by \( \fvertices \) and \( \rvertices \), are associated with valid states. The edges in the forward tree, \( \fedges \subseteq \fvertices \times \fvertices \), represent valid connections between states, while the edges in the reverse tree, \( \redges \subseteq \rvertices \times \rvertices \), can lead through invalid regions of the problem domain. An edge consists of a parent state, \( \sparent \), and a child state, \( \schild \), and is denoted as \( (\sparent, \schild) \).

Let \( \Rposinfty \) denote the union of the nonnegative real numbers with infinity. The function \( \ghatstandalone \colon \sspace \to \Rposinfty \) represents an admissible heuristic of the cost-to-come from the start to a state. The function \( \gtstandalone \colon \sspace \to \Rposinfty \) represents the cost-to-come from the start state to a state through the current forward tree. This cost is taken to be infinite for any state without an associated vertex in the forward tree.

The function \( \hhatstandalone \colon \sspace \to \Rposinfty \) represents an admissible heuristic of the cost-to-go from a state to a goal. The function \( \fhatstandalone \colon \sspace \to \Rposinfty \) represents an admissible estimate of the cost of a path from the start to a goal constrained to go through a state, e.g., \( \fhat{\sany} \defeq \ghat{\sany} + \hhat{\sany} \). This estimate defines the informed set of states that could provide a better solution, \( \sinformed \defeq \set{ \sany \in \sspace \given \fhat{\sany} \leq c_{\mathrm{current}}} \), where \( c_{\mathrm{current}} \) is the current solution cost. The function \( \ctruestandalone \colon \sspace \times \sspace \to \Rposinfty \) denotes the true edge cost between two states. The function \( \chatstandalone \colon \sspace \times \sspace \to \Rposinfty \) is an admissible estimate of this cost.

Let \( A \) be a set and let \( B \), \( C \) be subsets of \( A \). The notation \( B \overset{+}{\leftarrow} C \) is used for \( B \leftarrow B \cup C \) and \( B \overset{-}{\leftarrow} C \) for \( B \leftarrow B \setminus C \). The number of states sampled per batch is denoted by \( m \).

\begin{algorithm}[t]
  \caption{\small Concept of BIT* {\color{mpsred}\bfseries with changes for AIT*}}%
  \label{alg:conceptual}
  \SetInd{0.0em}{0.4em}
  \SetVlineSkip{0.0em}
  \tt\footnotesize
  initialize search, queue, and approximation\;
  {\color{mpsred}\textbf{update the heuristic}}\;
  \Repeat{\normalfont\tt stopped}{
      get and remove the best edge from the queue\;
      \eIf{\normalfont\tt the edge can possibly improve the solution}{
        \eIf{\normalfont\tt the edge is valid}{
          compute the true cost of the edge\;
          update the search tree with the edge\;
        }{
          {\color{mpsred}\textbf{update the heuristic}}\;
        }
      }{
        update the approximation\;
        {\color{mpsred}\textbf{update the heuristic}}\;
        populate the queue\;
      }
  }
\end{algorithm}%

\subsection{Approximation}%
\label{sec:approximation}

\aitstar{} samples batches of states to build a discrete approximation of the state space. It uses informed sampling to focus its approximation to the region of the state space that can possibly improve the current solution (Alg.~\ref{alg:technical}, line~\ref{alg:technical:line:sampling}).

States that are within a radius, \( r \), of each other are treated as neighbors (Alg.~\ref{alg:neighbors}, line~\ref{alg:neighbors:line:rgg}). Graph complexity is limited as states are sampled by decreasing this radius as in~\citep{karaman2011}, using the measure of the informed set, as in~\citep{gammell2018},
\vspace*{-0.7em}
\begin{align*}
  r(q) \defeq \eta {\left( 2 \left( 1 + \frac{1}{n} \right) \left( \frac{\lebesgue{\sinformed}}{\zeta_{n}} \right) \left( \frac{\log\left( q \right)}{q} \right) \right)}^{\frac{1}{n}},
\end{align*}
\vspace*{-0.2em}%
where \( q \) is the number of sampled states in the informed set, \( \eta > 1 \) is a tuning parameter, and \( \lambda(\sinformed) \) and \( \zeta_{n} \) are the Lebesgue measures of the informed set and an \( n \)-dimensional unit ball, respectively. Faster-decreasing radii are provided in~\citep{janson2015,janson2018} but are not used in \aitstar{} for fairer comparison to existing algorithms as they are presented in the literature.

\aitstar{} always includes the existing connections in both the forward and the reverse search trees in its approximation (Alg.~\ref{alg:neighbors}, lines~\ref{alg:neighbors:line:rggextbegin},~\ref{alg:neighbors:line:rggextend}), and removes invalid edges, regardless of the distance (Alg.~\ref{alg:updateheuristic}, line~\ref{alg:updateheuristic:line:blacklist}; Alg.~\ref{alg:neighbors}, line~\ref{alg:neighbors:line:blacklist}).

\begin{algorithm}[t]
  \caption{BIT* {\color{mpsred}\bfseries with changes for AIT*}}%
  \label{alg:technical}
  \SetInd{0.0em}{0.4em}
  \SetKwFunction{sample}{sample}
  \tt\footnotesize
  \( \fvertices \sset \sstart \); \( \fedges \sset \emptyset \); \( \ftree \sset (\fvertices, \fedges) \); \( c_{\mathrm{current}} \sset \infty \)\;
  \( \ssampled \sset \sgoals \cup \{ \sstart \}  \); \( \fqueue \sset \mpsexpand{\sstart} \);
  
  {\color{mpsred}\bfseries\(\updateheuristic{}\)}\;\label{alg:technical:line:initialize-heuristic}
  \Repeat{\normalfont\tt stopped}{
      \( (\sparent, \schild) \sset \popbest{\fqueue} \)\;
      \eIf{\( \gt{\sparent} + \chat{\sparent}{\schild} + \hhat{\schild} < c_{\mathrm{current}} \)}{\label{alg:technical:line:can-edge-possibly-improve-solution}
        \uIf{\( (\sparent, \schild) \in \fedges \)}{
          \( \fqueue \ssetinsert \mpsexpand{\schild}\)\;\label{alg:technical:line:freebie}
        }
        \ElseIf{\( \gt{\sparent} + \chat{\sparent}{\schild} < \gt{\schild} \)}{\label{alg:technical:line:can-edge-possibly-improve-tree}
          \eIf{\normalfont\(\isvalid{(\sparent, \schild)}\)}{\label{alg:technical:line:collision-detection}
            \If{\( \gt{\sparent} + \ctrue{\sparent}{\schild} + \hhat{\schild} < c_{\mathrm{current}} \)}{\label{alg:technical:line:can-edge-actually-improve-solution}
              \If{\( \gt{\sparent} + \ctrue{\sparent}{\schild} < \gt{\schild} \)}{\label{alg:technical:line:can-edge-actually-improve-tree}
                \eIf{\( \schild \not\in \fvertices \)}{
                  \( \fvertices \ssetinsert \schild \)\;\label{alg:technical:line:add-state-to-tree}
                }{
                  \( \fedges \ssetremove (\mpsfparent{\schild}, \schild) \)\;\label{alg:technical:line:rewiring}
                }
                \( \fedges \ssetinsert (\sparent, \schild) \)\;\label{alg:technical:line:add-edge-to-tree}
                \( \fqueue \ssetinsert \mpsexpand{\schild} \)\;\label{alg:technical:line:expand-child-state}
                \( c_{\mathrm{current}} \sset \min_{\sgoal \in \sgoals}\left\{ \gt{\sgoal} \right\} \)\;\label{alg:technical:line:update-solution-cost}
              }
            }
          }{
            {\color{mpsred}\bfseries\(\updateheuristic{\left\{ ( \sparent, \schild ) \right\}} \)}\;\label{alg:technical:line:update-heuristic}
          }
        }
      }{
        \( \ssampled \ssetinsert \mpssample{m, c_{\mathrm{current}}} \)\;\label{alg:technical:line:sampling}
        \( \fqueue \sset \mpsexpand{ \sstart} \)\;
        {\color{mpsred}\bfseries\(\updateheuristic{}\)}\;\label{alg:technical:line:reinitialize-heuristic}
      }
  }
\end{algorithm}%
%
\begin{algorithm}[t]
  \caption{\( \mpsexpand{\sany} \)}%
  \label{alg:expand}
  \SetInd{0.0em}{0.4em}
  \footnotesize
  \( E_{\mathrm{out}} \sset \emptyset \)\;
  \For{\normalfont\textbf{all} \( \snum{i} \in \mpsneighbors{\sany} \)}{
    \( E_{\mathrm{out}} \ssetinsert ( \sany, \snum{i} ) \)\;
  }
  \Return{ \( E_{\mathrm{out}} \) }
\end{algorithm}%
%
\begin{algorithm}[th!]
  \caption{\( \mpsneighbors{\sany} \)}%
  \label{alg:neighbors}
  \SetInd{0.0em}{0.4em}
  \SetKwFunction{fchildren}{children\({}_{\mathcal{F}}\)\hspace*{-0.2em}}
  \SetKwFunction{rchildren}{children\({}_{\mathcal{R}}\)\hspace*{-0.2em}}
  \footnotesize
  \( V_{\mathrm{neighbors}} \sset \set{ \snum{i} \in \ssampled \given \norm{\sany - \snum{i}} \leq r\left( q \right) } \)\;\label{alg:neighbors:line:rgg}
  \( V_{\mathrm{neighbors}} \ssetinsert \mpsfparent{\sany}\); {\color{mpsred}\( V_{\mathrm{neighbors}} \ssetinsert \mpsrparent{\sany} \)}\;\label{alg:neighbors:line:rggextbegin}
  \( V_{\mathrm{neighbors}} \ssetinsert \mpsfchildren{\sany} \); {\color{mpsred}\( V_{\mathrm{neighbors}} \ssetinsert \mpsrchildren{\sany} \)}\;\label{alg:neighbors:line:rggextend}
  {\color{mpsred}\bfseries\For{\normalfont\textbf{all} \( \snum{i} \in V_{\mathrm{neighbors}} \)}{
    \If{\( ( \sany, \snum{i} ) \in \edgeblacklist \) \normalfont\textbf{or} \( ( \snum{i}, \sany ) \in \edgeblacklist \)}{
      \( V_{\mathrm{neighbors}} \ssetremove \snum{i} \)\;\label{alg:neighbors:line:blacklist}
    }
  }}
  \Return{ \( V_{\mathrm{neighbors}} \) }
\end{algorithm}%
%

\subsection{Reverse Search}%
\label{sec:reverse-search}

\aitstar{} estimates a heuristic specific to the current approximation by performing a lazy reverse search with LPA* (Alg.~\ref{alg:updateheuristic} and Alg.~\ref{alg:updatestate}). This search uses a vertex-queue, \( \rqueue \), which sorts states according to a lexicographical key,
\begin{align*}
  \texttt{key}_{\mathrm{R}}(\sany) \defeq &\left(\min\left\{ \hhatcon{\sany}, \hhatexp{\sany} \right\} + \ghat{\sany},\right.\\ &\phantom{\Big(}\left.\min\left\{ \hhatcon{\sany}, \hhatexp{\sany} \right\} \right),
\end{align*}
\vspace*{-0.08em}%
where \( \hhatcon{\sany} \) denotes the cost-to-go of \( \sany \) when it was last connected to the reverse tree and \( \hhatexp{\sany} \) denotes the cost-to-go of \( \sany \) when it was last expanded in the reverse search. These are the \( g \) and \( v \) values in a forward LPA* search~\citep{aine2016}.

If Algorithm~\ref{alg:updateheuristic} is called without an argument (Alg.~\ref{alg:technical}, lines~\ref{alg:technical:line:initialize-heuristic},~\ref{alg:technical:line:reinitialize-heuristic}), it sets the \( \hhatconstandalone \) and \( \hhatexpstandalone \) values of all states except the goals to infinity and inserts the goal states into the queue (Alg.~\ref{alg:updateheuristic}, lines~\ref{alg:updateheuristic:line:restart-begin}--\ref{alg:updateheuristic:line:restart-end}). This restarts LPA*, which is more efficient than repairing the search when large changes in the graph are expected~\citep{likhachev2005,likhachev2008,aine2016}. LPA*'s initial search is equivalent to A* and results in a consistent and admissible estimate of the heuristic in the current approximation.

If Algorithm~\ref{alg:updateheuristic} is called with an invalid edge (Alg.~\ref{alg:technical}, line~\ref{alg:technical:line:update-heuristic}), then it adds the edge to the set of invalid edges, \( \edgeblacklist \), and updates the cost-to-go of the parent state (Alg.~\ref{alg:updateheuristic}, lines~\ref{alg:updateheuristic:line:blacklist},~\ref{alg:updateheuristic:line:update-parent-state}). LPA* then repairs the reverse search tree and increases the cost-to-go values, \( \hhatconstandalone \), as necessary. This results in an updated heuristic which is still admissible for the current approximation and can be used by the forward search. Full details of LPA* are available in~\citep{koenig2004,likhachev2005,aine2016}.

\subsection{Forward Search}%
\label{sec:forward-search}

The forward search of \aitstar{} uses an edge-queue, \( \fqueue \), which sorts edges according to a  lexicographical key,
\begin{align*}
  \texttt{key}_{\mathrm{F}}(\sparent, \schild) \defeq &\left( \gt{\sparent} + \chat{\sparent}{\schild} + \hhat{\schild},\right.\\ &\left.\phantom{\Big(}\gt{\sparent} + \chat{\sparent}{\schild},\; \gt{\sparent} \right),
\end{align*}
where the cost-to-go values from the reverse search are used as heuristic for the forward search, i.e., \( \hhat{\sany} \defeq \hhatcon{\sany} \).

\begin{algorithm}[t]
  \caption{\color{mpsred}\bfseries\( \updateheuristic{ ( \sparent, \schild ) }\)}%
  \label{alg:updateheuristic}
  \SetInd{0.0em}{0.4em}
  \footnotesize
  \eIf{\( ( \sparent, \schild ) = \emptyset \)}{\label{alg:updateheuristic:line:restart-begin}
    \For{\normalfont\textbf{all} \( \sany \in \rvertices \)}{
      \( \hhatcon{\sany} \sset \infty \);
      \( \hhatexp{\sany} \sset \infty \)\;
    }
    \For{\normalfont\textbf{all} \( \sgoal \in \sgoals \)}{
      \( \hhatcon{\sgoal} \sset 0 \)\;
      \( \rqueue \sset \sgoal \)\;
    }\label{alg:updateheuristic:line:restart-end}
  }{
    \( \edgeblacklist \ssetinsert ( \sparent, \schild ) \)\;\label{alg:updateheuristic:line:blacklist}
    \( \mpsupdatestate{\sparent} \)\;\label{alg:updateheuristic:line:update-parent-state}
  }
    \While{\normalfont\( \min_{\sany \in \rqueue} \left\{ \texttt{key}_{\mathrm{R}}(\sany) \right\} < \texttt{key}_{\mathrm{R}}(\sstart) \)\\\normalfont\textbf{ or } \( \hhatexp{\sstart} < \hhatcon{\sstart} \) \\\normalfont\textbf{ or } \( \fqueue \) contains an edge with an unprocessed vertex}{
      \( \sany \sset \popbest{\rqueue} \)\;
      \eIf{\( \hhatcon{\sany} < \hhatexp{\sany} \)}{
        \( \hhatexp{\sany} \sset \hhatcon{\sany} \)\;
      }{
        \( \hhatexp{\sany} \sset \infty \)\;
        \( \mpsupdatestate{\sany}\)\;
      }
      \For{\normalfont\textbf{all }\( \snum{i} \in \mpsneighbors{\sany} \)}{
        \( \mpsupdatestate{\snum{i}} \)\;
      }
  }
\end{algorithm}%

%
\begin{algorithm}[t]
  \caption{\color{mpsred}\bfseries \( \mpsupdatestate{\sany} \)}%
  \label{alg:updatestate}
  \SetInd{0.0em}{0.4em}
  \footnotesize
  \If{\( \sany \neq \sstart \)}{
    \( \sparent \sset \mathop{\arg\,\min}_{\snum{i} \in \mpsneighbors{\sany}} \left\{ \hhatexp{\snum{i}} + \chat{\snum{i}}{\sany} \right\} \)\;
    \( \hhatcon{\sany} \sset \hhatexp{\sparent} + \chat{\sparent}{\sany} \)\;
    \uIf{\( \hhatcon{\sany} \neq \hhatexp{\sany} \)}{
      \If{\( \sany \not\in \rqueue \)}{
        \( \rqueue \ssetinsert \sany \)
      }
    }
    \ElseIf{\( \sany \in \rqueue \)}{
      \( \rqueue \ssetremove \sany \)
    }
  }
\end{algorithm}
%

An iteration begins by getting the best edge from the queue and checking whether it can possibly improve the current solution (Alg.~\ref{alg:technical}, line~\ref{alg:technical:line:can-edge-possibly-improve-solution}). If it can and is already part of the forward tree, its child state is expanded (Alg.~\ref{alg:technical}, line~\ref{alg:technical:line:freebie}).

If it is not in the forward tree but can possibly improve it (Alg.~\ref{alg:technical}, line~\ref{alg:technical:line:can-edge-possibly-improve-tree}), then the edge is checked for validity (Alg.~\ref{alg:technical}, line~\ref{alg:technical:line:collision-detection}). If the edge is invalid, then the heuristic is updated by the reverse search (Alg.~\ref{alg:technical}, line~\ref{alg:technical:line:update-heuristic}; Alg~\ref{alg:updateheuristic}). If the edge is valid, then it is completely evaluated. The search then checks whether it can actually improve the current solution and the forward tree (Alg.~\ref{alg:technical}, lines~\ref{alg:technical:line:can-edge-actually-improve-solution},~\ref{alg:technical:line:can-edge-actually-improve-tree}).

The child state of a new edge that can improve the current solution and the forward tree is added to the tree if it is not already in the tree (Alg.~\ref{alg:technical}, line~\ref{alg:technical:line:add-state-to-tree}). If it is, then the new edge is a rewiring and the old edge is removed (Alg.~\ref{alg:technical}, line~\ref{alg:technical:line:rewiring}). The new edge is added to the tree and its child state is expanded in both cases (Alg.~\ref{alg:technical}, lines~\ref{alg:technical:line:add-edge-to-tree},~\ref{alg:technical:line:expand-child-state}).

The iteration finishes by updating the current solution cost (Alg.~\ref{alg:technical}, line~\ref{alg:technical:line:update-solution-cost}). In practice this is done efficiently by only checking the goal states in the forward search tree.

If the forward search processes an edge that can not possibly improve the current solution, then a new search on a refined approximation is started (Alg.~\ref{alg:technical}, lines~\ref{alg:technical:line:can-edge-possibly-improve-solution},~\ref{alg:technical:line:sampling}--\ref{alg:technical:line:reinitialize-heuristic}).

\section{Experimental Results}%
\label{sec:experimental-results}

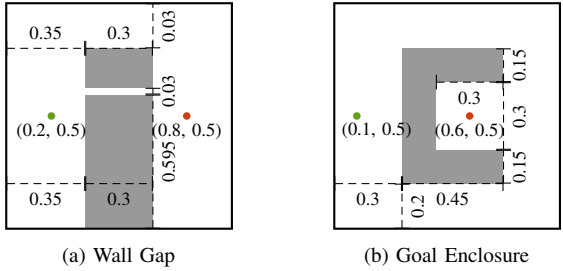
\begin{figure}[t]
  \begin{minipage}[b]{0.495\columnwidth}
    \centering%
    \begin{tikzpicture}[scale = 3.0]
\draw [fill = black!40, draw = none] (-0.15, -0.5) rectangle (0.15, 0.3);

\draw [draw = black, fill = none, thick] (-0.5, -0.5) rectangle (0.5, 0.5);

\draw [fill = white, draw = none] (-0.16, 0.095) rectangle (0.16, 0.125);

\node (wallgapstart) [fill = mpsgreen, inner sep = 0mm, circle, minimum size = 1mm] at (-0.3, 0.0) {};
\node [below = 0pt of wallgapstart, inner sep = 1pt] {\scriptsize (0.2, 0.5)};

\node (wallgapgoal) [fill = mpsred, inner sep = 0mm, circle, minimum size = 1mm] at (0.3, 0.0) {};
\node [below = 0pt of wallgapgoal, inner sep = 1pt] {\scriptsize (0.8, 0.5)};

\draw [|-|, densely dashed] (-0.15, -0.3) -- node [below, midway] {\scriptsize 0.3} (0.15, -0.3);
\draw [|-|, densely dashed] (-0.5, -0.3) -- node [below, midway] {\scriptsize 0.35} (-0.15, -0.3);
\draw [|-|, densely dashed] (0.15, -0.5) -- node [below, sloped, midway] {\scriptsize 0.595} (0.15, 0.095);
\draw [|-|, densely dashed] (0.15, 0.095) -- node [below, sloped, midway] {\scriptsize 0.03} (0.15, 0.125);
\draw [|-|, densely dashed] (0.15, 0.3) -- node [below, sloped, midway] {\scriptsize 0.03} (0.15, 0.5);
\draw [|-|, densely dashed] (-0.15, 0.3) -- node [above, midway] {\scriptsize 0.3} (0.15, 0.3);
\draw [|-|, densely dashed] (-0.5, 0.3) -- node [above, midway] {\scriptsize 0.35} (-0.15, 0.3);
\end{tikzpicture}%
%
    \subcaption{Wall Gap}%
    \label{fig:experiment-wall-gap}
  \end{minipage}%
  \hfill%
  \begin{minipage}[b]{0.495\columnwidth}
    \centering%
    \begin{tikzpicture}[scale = 3.0]
\draw [draw = black, fill = none, thick] (-0.5, -0.5) rectangle (0.5, 0.5);

\draw [fill = black!40, draw = none] (-0.2, -0.3) rectangle (0.4, 0.3);

\draw [fill = white, draw = none] (-0.05, -0.15) rectangle (0.251, 0.15);
\draw [fill = white, draw = none] (0.25, -0.31) rectangle (0.41, 0.31);

\node (goalenclosurestart) [fill = mpsgreen, inner sep = 0mm, circle, minimum size = 1mm] at (-0.4, 0.0) {};
\node [below right = 0pt and -8pt of goalenclosurestart, inner sep = 1pt] {\scriptsize (0.1, 0.5)};

\node (goalenclosuregoal) [fill = mpsred, inner sep = 0mm, circle, minimum size = 1mm] at (0.1, 0.0) {};
\node [below = 0pt of goalenclosuregoal, inner sep = 1pt] {\scriptsize (0.6, 0.5)};

\draw [|-|, densely dashed] (-0.2, -0.3) -- node [below, midway] {\scriptsize 0.45} (0.25, -0.3);
\draw [|-|, densely dashed] (-0.05, 0.15) -- node [below, midway] {\scriptsize 0.3} (0.25, 0.15);
\draw [|-|, densely dashed] (0.25, 0.15) -- node [sloped, below, midway] {\scriptsize 0.15} (0.25, 0.3);
\draw [|-|, densely dashed] (0.25, -0.15) -- node [sloped, below, midway] {\scriptsize 0.3} (0.25, 0.15);
\draw [|-|, densely dashed] (0.25, -0.3) -- node [sloped, below, midway] {\scriptsize 0.15} (0.25, -0.15);
\draw [|-|, densely dashed] (-0.2, -0.5) -- node [sloped, below, midway] {\scriptsize 0.2} (-0.2, -0.3);
\draw [|-|, densely dashed] (-0.5, -0.3) -- node [sloped, below, midway] {\scriptsize 0.3} (-0.2, -0.3);

\end{tikzpicture}%
%
    \subcaption{Goal Enclosure}%
    \label{fig:experiment-goal-enclosure}
  \end{minipage}
  \caption{Two dimensional illustrations of the experiments on which the planners were tested. The green (\protect\tikz{\protect\node[circle, fill=white, inner sep = 0.5mm] {}; \protect\node[circle, fill=mpsgreen, inner sep = 0.4mm] {};}) and red (\protect\tikz{\protect\node[circle, fill=white, inner sep = 0.5mm] {}; \protect\node[circle, fill=mpsred, inner sep = 0.4mm] {};}) dots represent the positions of the start and goal states, respectively. Grey regions represent invalid states. Each state space dimension was bounded to the interval [0, 1], which is illustrated by the black bounding boxes.}%
  \label{fig:experiments}
\end{figure}%

\aitstar{} was compared against the Open Motion Planning Library (OMPL)~\citep{sucan2012} implementations of RRT-Connect, RRT*, RRT\({}^{\#}\)~\citep{arslan2013}, and BIT* on simulated problems\footnote{The performances were measured with OMPL v1.4.1 on a laptop with 16~GB of RAM and an Intel i7-4910MQ processor running Ubuntu 18.04.}. RRT* and RRT\({}^{\#}\) used a goal bias of 5\% and RRT\({}^{\#}\) used rejection sampling. All RRT-based algorithms used maximum edge lengths of 0.5, 1.25, and 3.0 in \( \mathbb{R}^{4}, \mathbb{R}^{8} \), and \( \mathbb{R}^{16} \), respectively. BIT* and \aitstar{} sampled 100 states per batch regardless of dimension and used the Euclidean norm for all \textit{a priori} heuristics. The RGG constant \( \eta \) was 1.001 for all planners.

\subsection{Abstract Problems}%
\label{sec:abstract-problems}

The planners were tested on two abstract problems with different obstacle configurations in \( \mathbb{R}^{4}, \mathbb{R}^{8} \), and \( \mathbb{R}^{16} \) (Fig.~\ref{fig:experiments}). Each planner was run 100 times with different random seeds on each instantiation. Planners were given one second to solve problems in \( \mathbb{R}^{4} \), ten seconds in \( \mathbb{R}^{8} \), and 100 seconds in \( \mathbb{R}^{16} \). The collision detection resolution was set to \( 10^{-6} \) to make evaluating edge costs computationally expensive. The optimization objective was path length. Figure~\ref{fig:results} shows the achieved preformances of all tested planners on all problems.

One abstract problem consisted of a wall with a narrow gap, such that in all dimensions only two homotopy classes exist (Fig.~\ref{fig:experiment-wall-gap}). This shows \aitstar{}'s performance on a problem containing a hard-to-find optimal homotopy class (Figs.~\ref{fig:results-wall-gap-r4}--\subref{fig:results-wall-gap-r16}).

The other abstract problem consisted of a hollow, axis-aligned hyperrectangle enclosing the goal state configured such that even in higher dimensions the goal can only be reached through the face of the hyperrectangle farthest from the start state (Fig.~\ref{fig:experiment-goal-enclosure}). This problem is challenging for \aitstar{} because there are many invalid edges close to the root of the reverse search tree which means that often large parts of it must be repaired (Figs.~\ref{fig:results-goal-enclosure-r4}--\subref{fig:results-goal-enclosure-r16}).

\subsection{Planning for Axel}%
\label{sec:planning-for-axel}

The benefits of \aitstar{}'s asymmetric bidirectional search were also tested on simulated planning problems for NASA/JPL-Caltech's Axel Rover System (Fig.~\ref{fig:teaser}), which is specialized for challenging terrain. These problems require sequences of \( \SEthree \) poses settled on the surface manifold of the terrain. This makes edge evaluations expensive, as every state along an edge has to be projected onto the manifold.

\begin{figure*}[t]
  \begin{minipage}[b]{0.38\textwidth}
    \input{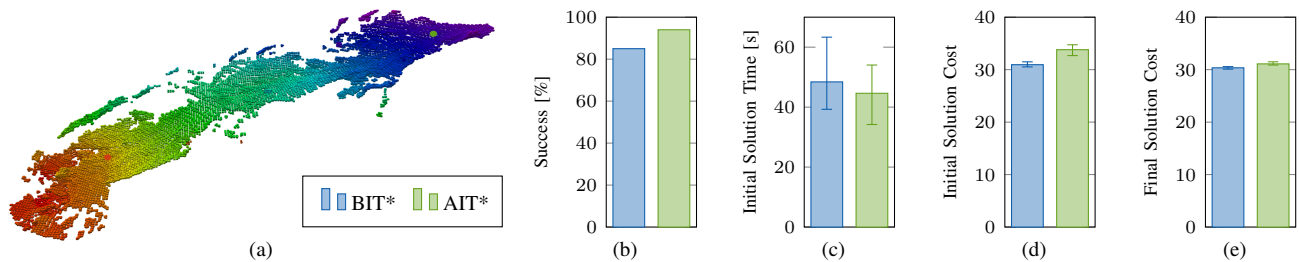}%
  \vspace*{-0.7em}%
  \subcaption{}%
  \label{fig:results-axel-map}
  \end{minipage}\hfill%
  \begin{minipage}[b]{0.15\textwidth}
      \begin{tikzpicture}
    \begin{axis}[
      x = 0.6cm,
      width = \textwidth,
      height = 1.64\textwidth,
      ymin = 0,
      ymax = 100,
      bar width = 1.2em,
      enlarge x limits ={abs=0.3cm},
      symbolic x coords={bitstar, aibitstar},
      xtick = {\empty},
      ylabel = {Success [\%]},
      yticklabel style={font=\scriptsize},
      ylabel style={font=\scriptsize, text depth=-0.0em, text height=0.5em}
      ]
      \addplot [ybar, mpsblue, fill = mpsblue!40,
                error bars/.cd, y dir=both, y explicit] coordinates
      {
        (bitstar, 85)
      };
      \addplot [ybar, mpsgreen, fill = mpsgreen!40,
                error bars/.cd, y dir=both, y explicit] plot coordinates
      {
        (aibitstar, 94)
      };
    \end{axis}
  \end{tikzpicture}%
  \vspace*{-0.7em}%
  \subcaption{}%
  \label{fig:results-axel-success}
  \end{minipage}\hspace*{0.3em}%
  \begin{minipage}[b]{0.15\textwidth}
      \begin{tikzpicture}
    \begin{axis}[
      x = 0.6cm,
      width = \textwidth,
      height = 1.64\textwidth,
      ymin = 0,
      ymax = 70,
      bar width = 1.2em,
      enlarge x limits ={abs=0.3cm},
      symbolic x coords={bitstar, aibitstar},
      xtick = {\empty},
      ylabel = {Initial Solution Time [s]},
      yticklabel style={font=\scriptsize},
      ylabel style={font=\scriptsize, text depth=-0.0em, text height=0.5em}
      ]
      \addplot [ybar, mpsblue, fill = mpsblue!40,
                error bars/.cd, y dir=both, y explicit] coordinates
      {
        (bitstar, 48.4125) += (0, 14.9222) -= (0, 9.1348)
      };
      \addplot [ybar, mpsgreen, fill = mpsgreen!40,
                error bars/.cd, y dir=both, y explicit] plot coordinates
      {
        (aibitstar, 44.6175) += (0, 9.4210) -= (0, 10.4108)
      };
    \end{axis}
  \end{tikzpicture}%
  \vspace*{-0.7em}%
  \subcaption{}%
  \label{fig:results-axel-initial-solution-time}
  \end{minipage}%
  \begin{minipage}[b]{0.15\textwidth}
      \begin{tikzpicture}
    \begin{axis}[
      x = 0.6cm,
      width = \textwidth,
      height = 1.64\textwidth,
      ymin = 0,
      ymax = 40,
      bar width = 1.2em,
      enlarge x limits={abs=0.3cm},
      symbolic x coords={bitstar, aibitstar},
      xtick = {\empty},
      ylabel = {Initial Solution Cost},
      yticklabel style={font=\scriptsize},
      ylabel style={font=\scriptsize, text depth=-0.0em, text height=0.5em}
      ]
      \addplot [ybar, mpsblue, fill = mpsblue!40,
                error bars/.cd, y dir=both, y explicit] coordinates
      {
        (bitstar, 30.9463) += (0, 0.5463) -= (0, 0.4190)
      };
      \addplot [ybar, mpsgreen, fill = mpsgreen!40,
                error bars/.cd, y dir=both, y explicit] plot coordinates
      {
        (aibitstar, 33.8055) += (0, 0.9404) -= (0, 1.1230)
      };
    \end{axis}
  \end{tikzpicture}%
  \vspace*{-0.7em}%
  \subcaption{}%
  \label{fig:results-axel-initial-solution-cost}
  \end{minipage}%
  \begin{minipage}[b]{0.15\textwidth}
      \begin{tikzpicture}
    \begin{axis}[
      x = 0.6cm,
      width = \textwidth,
      height = 1.64\textwidth,
      ymin = 0,
      ymax = 40,
      bar width = 1.2em,
      enlarge x limits={abs=0.3cm},
      symbolic x coords={bitstar, aibitstar},
      xtick = {\empty},
      ylabel = {Final Solution Cost},
      yticklabel style={font=\scriptsize},
      ylabel style={font=\scriptsize, text depth=-0.0em, text height=0.5em}
      ]
      \addplot [ybar, mpsblue, fill = mpsblue!40,
                error bars/.cd, y dir=both, y explicit] coordinates
      {
        (bitstar, 30.3507) += (0, 0.2279) -= (0, 0.2453)
      };
      \addplot [ybar, mpsgreen, fill = mpsgreen!40,
                error bars/.cd, y dir=both, y explicit] plot coordinates
      {
        (aibitstar, 31.1006) += (0, 0.3981) -= (0, 0.22061)
      };
    \end{axis}
  \end{tikzpicture}%
  \vspace*{-0.7em}%
  \subcaption{}%
  \label{fig:results-axel-final-solution-cost}
  \end{minipage}
  \vspace*{-0.0em}%
  \caption{Results from 100 trials of BIT* and AIT* on a problem for NASA/JPL-Caltech's Axel. The challenge was to plan down a steep slope (\subref{fig:results-axel-map}) from the green dot (top right) to the red dot (bottom left), through two narrow passages. The map is colored by elevation. The plot~(\subref{fig:results-axel-success}) shows the achieved success rates after running for 100 seconds. The plots~(\subref{fig:results-axel-initial-solution-time}),~(\subref{fig:results-axel-initial-solution-cost}), and~(\subref{fig:results-axel-final-solution-cost}) show the medians with nonparametric 99\% confidence intervals of the initial solution times, the initial solution costs, and the final solution costs, respectively. Unsuccessful runs were taken as infinite costs. AIT* achieves a higher success rate and slighly faster initial solution times but higher initial solution costs than BIT*. Both planners achieve similar final solution costs.}%
  \label{fig:results-axel}
  \vspace*{-0.8em}
\end{figure*}

BIT* and \aitstar{} were run 100 times to plan a path down a steep slope with a line-of-sight distance of 30.97 meters between the start and goal positions (Fig.~\ref{fig:results-axel-map}). The linear and angular collision detection resolutions were set to 2 cm and 0.1 rad, respectively. BIT* and \aitstar{} optimized for path length and roll. Figure~\ref{fig:results-axel} shows the achieved performances.

\begin{figure*}[t]
  \begin{minipage}[b]{0.33\textwidth}
    \vspace*{0.5em}%
    \input{figures/results/synthetic/wall_gap_r4.tikz}%
    \vspace*{-0.5em}%
    \subcaption{Wall Gap in \( \mathbb{R}^{4} \)}\label{fig:results-wall-gap-r4}
  \end{minipage}%
  \begin{minipage}[b]{0.33\textwidth}
    \vspace*{0.5em}%
    \input{figures/results/synthetic/wall_gap_r8.tikz}%
    \vspace*{-0.5em}%
    \subcaption{Wall Gap in \( \mathbb{R}^{8} \)}\label{fig:results-wall-gap-r8}
  \end{minipage}%
  \begin{minipage}[b]{0.33\textwidth}
    \vspace*{0.5em}%
    \input{figures/results/synthetic/wall_gap_r16.tikz}%
    \vspace*{-0.5em}%
    \subcaption{Wall Gap in \( \mathbb{R}^{16} \)}\label{fig:results-wall-gap-r16}
  \end{minipage}\\
  \begin{minipage}[b]{0.33\textwidth}
    \vspace*{0.5em}%
    \input{figures/results/synthetic/goal_enclosure_r4.tikz}%
    \vspace*{-0.5em}%
    \subcaption{Goal Enclosure in \( \mathbb{R}^{4} \)}\label{fig:results-goal-enclosure-r4}
  \end{minipage}%
  \begin{minipage}[b]{0.33\textwidth}
    \vspace*{0.5em}%
    \input{figures/results/synthetic/goal_enclosure_r8.tikz}%
    \vspace*{-0.5em}%
    \subcaption{Goal Enclosure in \( \mathbb{R}^{8} \)}\label{fig:results-goal-enclosure-r8}
  \end{minipage}%
  \begin{minipage}[b]{0.33\textwidth}
    \vspace*{0.5em}%
    \input{figures/results/synthetic/goal_enclosure_r16.tikz}%
    \vspace*{-0.5em}%
    \subcaption{Goal Enclosure in \( \mathbb{R}^{16} \)}\label{fig:results-goal-enclosure-r16}
  \end{minipage}\\
  \begin{minipage}[b]{\textwidth}
    \centering%
    \vspace*{0.5em}%
    \begin{tikzpicture}
\begin{axis} [
  width=\textwidth,
  height=0.5\textwidth,
  unbounded coords=jump,
  xtick align=inside,
  ytick align=inside,
  anchor=north,
  hide axis,
  xmajorgrids,
  ymajorgrids,
  major grid style={densely dotted, black!20},
  xmin=0,
  xmax=10,
  ymin=0,
  ymax=10,
  xlabel style={font=\footnotesize},
  xticklabel style={font=\footnotesize},
  ylabel style={font=\footnotesize},
  yticklabel style={font=\footnotesize},
  legend style={font=\footnotesize, anchor=south, legend cell align=left, legend columns=-1, at={(axis cs:5, 6)}}
]
\addlegendimage{black, line width = 1.0pt, mark size=1.0pt, mark=square*}
\addlegendentry{RRT-Connect}
\addlegendimage{mpsred, line width = 1.0pt, mark size=1.0pt, mark=square*}
\addlegendentry{RRT*}
\addlegendimage{mpsyellow, line width = 1.0pt, mark size=1.0pt, mark=square*}
\addlegendentry{RRT$^{\#}$}
\addlegendimage{mpsblue, line width = 1.0pt, mark size=1.0pt, mark=square*}
\addlegendentry{BIT*}
\addlegendimage{mpsgreen, line width = 1.0pt, mark size=1.0pt, mark=square*}
\addlegendentry{AIT*}
\end{axis}
\end{tikzpicture}
  \end{minipage}
  \caption{Planner performances on the abstract problems described in Section~\ref{sec:abstract-problems}. Results from the wall gap experiments are presented in plots~(\subref{fig:results-wall-gap-r4}),~(\subref{fig:results-wall-gap-r8}), and~(\subref{fig:results-wall-gap-r16}), and from the goal enclosure experiment in plots~(\subref{fig:results-goal-enclosure-r4}),~(\subref{fig:results-goal-enclosure-r8}), and~(\subref{fig:results-goal-enclosure-r16}). The squares in the cost plots show the median times and costs of the initial solutions with a nonparametric 99\% confidence interval. The lines show the median cost over time for almost-surely asymptotically optimal planners (unsuccessful runs were taken as infinite costs). Note that in plots~(\subref{fig:results-wall-gap-r16}),~(\subref{fig:results-goal-enclosure-r4}),~(\subref{fig:results-goal-enclosure-r8}), and~(\subref{fig:results-goal-enclosure-r16}) less than 50 trials of RRT* and RRT\( {}^{\#} \) were successful, so the median solution cost is infinite for these planners. AIT* finds initial solutions faster than RRT-Connect on four out of six problems, always faster than BIT* with the Euclidean heuristic, and around an order of magnitude faster than RRT* and RRT\( {}^{\#} \). The goal enclosure is a challenging problem for AIT*, because many states close to the goal are initially connected through invalid edges, which results in large updates of the reverse tree.}\vspace*{-1em}%
  \label{fig:results}
\end{figure*}

\section{Discussion \& Future Work}%
\label{sec:discussion-and-future-work}

\aitstar{} was designed for planning problems with expensive edge evaluations. These often occur when the search has to consider dynamic constraints (e.g., two-point boundary value problems) or complex robot and obstacle interactions (e.g., difficult collision detection) for each edge, as found on NASA/JPL-Caltech's Axel. In future work, Axel will consider tether-terrain interaction and physics-based stability checks based on the anchor history of the tether, which will further increase the edge evaluation cost.

These expensive edge evaluations were simulated in the abstract problems by increasing the collision detection resolution, providing a simple way to increase the edge evaluation cost and evaluate \aitstar{} on illustrative obstacle configurations.

The adaptive heuristic of \aitstar{} is less effective when the lazy reverse search connects many states through invalid edges, especially if these edges are near the root of the reverse search tree. This was illustrated with the goal enclosure experiment (Fig.~\ref{fig:experiment-goal-enclosure}). Future work could use sparse collision detection on the reverse search to mitigate this problem.

The reverse search of \aitstar{} could also be used to estimate the search effort instead of the solution cost. The forward search could then be replaced with an anytime search that explicitly tries to minimize the time to the next solution, similar to~\citep{thayer2012}, which could speed up initial solution times.

Another way to speed up initial solution times of AIT* would be to inflate the heuristic term in the key of the forward queue, as in Advanced BIT* (ABIT*)~\citep{strub2020}.

\section{Conclusion}%
\label{sec:conclusion}

Informed sampling-based algorithms use heuristic knowledge about a problem domain to improve their performance. Heuristics that are applicable to all problems in a domain are often simple to define and inexpensive to evaluate but seldom accurate for a specific problem instance. Problem-specific heuristics can be very accurate but the computational cost to estimate and/or evaluate them can often outweigh the improved search efficiency.

This paper presents \aitstar{}, an almost-surely asymptotically optimal sampling-based planner that simultaneously estimates and exploits an accurate heuristic specific to each problem instance. \aitstar{} uses an asymmetric bidirectional search to efficiently share information between the individual searches. The computationally inexpensive reverse search informs the expensive forward search by providing accurate heuristics specific to the current approximation of each problem instance. The forward search informs the reverse search by providing information about invalid edges, which results in ever more accurate heuristics. This is done efficiently by using LPA* as the reverse search algorithm.

This approach is promising for path planning problems with expensive edge evaluations, such as those posed by NASA/JPL-Caltech's Axel. \aitstar{} outperforms existing sampling-based algorithms on the tested abstract problems by finding an initial solution quickly and converging to the optimum in an anytime manner. These problems show the robustness of \aitstar{} with respect to expensive edge evaluations and encourage more thorough evaluations of states which could be used in more advanced optimization objectives.

Information on the OMPL implementation of AIT* is available at \texttt{\small\href{https://robotic-esp.com/code/}{https://robotic-esp.com/code/}}.

\bibliographystyle{IEEEtran}
\bibliography{tbdstar_bibliography}

\begin{thebibliography}{10}
\providecommand{\url}[1]{#1}
\csname url@samestyle\endcsname
\providecommand{\newblock}{\relax}
\providecommand{\bibinfo}[2]{#2}
\providecommand{\BIBentrySTDinterwordspacing}{\spaceskip=0pt\relax}
\providecommand{\BIBentryALTinterwordstretchfactor}{4}
\providecommand{\BIBentryALTinterwordspacing}{\spaceskip=\fontdimen2\font plus
\BIBentryALTinterwordstretchfactor\fontdimen3\font minus
  \fontdimen4\font\relax}
\providecommand{\BIBforeignlanguage}[2]{{%
\expandafter\ifx\csname l@#1\endcsname\relax
\typeout{** WARNING: IEEEtran.bst: No hyphenation pattern has been}%
\typeout{** loaded for the language `#1'. Using the pattern for}%
\typeout{** the default language instead.}%
\else
\language=\csname l@#1\endcsname
\fi
#2}}
\providecommand{\BIBdecl}{\relax}
\BIBdecl

\bibitem{kavraki1996}
L.~E. Kavraki, P.~{\v{S}}vestka, J.-C. Latombe, and M.~H. Overmars,
  ``Probabilistic roadmaps for path planning in high dimensional configuration
  spaces,'' \emph{IEEE Transactions on Robotics and Automation}, vol.~12,
  no.~4, pp. 566--580, 1996.

\bibitem{hart1968}
P.~E. Hart, N.~J. Nilsson, and B.~Raphael, ``A formal basis for the heuristic
  determination of minimum cost paths,'' \emph{{IEEE} Transactions on Systems
  Science and Cybernetics}, vol.~4, no.~2, pp. 100--107, 1968.

\bibitem{korf1997}
R.~E. Korf, ``Finding optimal solutions to {Rubik}'s cube using pattern
  databases,'' in \emph{Proceedings of the AAAI National Conference on
  Artificial Intelligence}, 1997, pp. 700--705.

\bibitem{culberson1998}
J.~C. Culberson and J.~Schauffer, ``Pattern databases,'' \emph{Computational
  Intelligence}, vol.~14, pp. 318--334, 1998.

\bibitem{felner2004}
A.~Felner, R.~E. Korf, and S.~Hanan, ``Additive pattern database heuristics,''
  \emph{Journal of Artificial Intelligence Research (JAIR)}, vol.~22, pp.
  279--318, 2004.

\bibitem{paden2017}
B.~Paden, V.~Varricchio, and E.~Frazzoli, ``Verification and synthesis of
  admissible heuristics for kinodynamic motion planning,'' \emph{IEEE Robotics
  and Automation Letters (RA-L)}, vol.~2, no.~2, pp. 648--655, 2017.

\bibitem{nesnas2012}
I.~A.~D. Nesnas, J.~B. Matthews, P.~Abad-Manterola, J.~W. Burdick, J.~A.
  Edlund, J.~C. Morrison, R.~D. Peters, M.~M. Tanner, R.~N. Miyake, B.~S.
  Solish \emph{et~al.}, ``{Axel} and {DuAxel} rovers for the sustainable
  exploration of extreme terrains,'' \emph{Journal of Field Robotics}, vol.~29,
  no.~4, pp. 663--685, 2012.

\bibitem{kleinbort2016}
M.~Kleinbort, O.~Salzman, and D.~Halperin, ``Collision detection or
  nearest-neighbor search? {On} the computational bottleneck in sampling-based
  motion planning,'' in \emph{Proceedings of the International Workshop on the
  Algorithmic Foundations of Robotics (WAFR)}, 2016.

\bibitem{bohlin2000}
R.~Bohlin and L.~E. Kavraki, ``Path planning using lazy {PRM},'' in
  \emph{Proceedings of the IEEE International Conference on Robotics and
  Automation (ICRA)}, 2000, pp. 521--528.

\bibitem{thayer2011}
J.~T. Thayer, A.~Dionne, and W.~Ruml, ``Learning inadmissible heuristics during
  search,'' in \emph{Twenty-First International Conference on Automated
  Planning and Scheduling}, 2011.

\bibitem{koenig2005}
S.~Koenig and M.~Likhachev, ``Adaptive {A}*,'' in \emph{In Proceedings of the
  International Converence on Autonomous Agents and Multiagent Systems
  (AAMAS)}, 2005, pp. 1311--1312.

\bibitem{koenig2006a}
S.~Koenig and M.~Likhachev, ``A new principle for incremental heuristic search:
  Theoretical results,'' in \emph{In Proceedings of the International
  Conference on Automated Planning and Scheduling (ICAPS)}, 2006.

\bibitem{kaindl1997}
H.~Kaindl and G.~Kainz, ``Bidirectional heuristic search reconsidered,''
  \emph{Journal of Artificial Intelligence Research (JAIR)}, vol.~7, no.~7, pp.
  283--317, 1997.

\bibitem{wilt2013}
C.~Wilt and W.~Ruml, ``Robust bidirectional search via heuristic improvement,''
  in \emph{Proceedings the Association for the Advancement of Artificial
  Intelligence Conference International Conference (AAAI)}, 2013.

\bibitem{kuffner2000}
J.~J. Kuffner~Jr. and S.~M. LaValle, ``{RRT}-{C}onnect: An efficient approach
  to single-query path planning,'' in \emph{Proceedings of the IEEE
  International Conference on Robotics and Automation (ICRA)}, 2000, pp.
  995--1001.

\bibitem{lavalle2001}
S.~M. LaValle and J.~J. Kuffner~Jr., ``Randomized kinodynamic planning,''
  \emph{The International Journal of Robotics Research (IJRR)}, vol.~20, no.~5,
  pp. 378--400, 2001.

\bibitem{gammell2018}
J.~D. Gammell, T.~D. Barfoot, and S.~S. Srinivasa, ``Informed sampling for
  asymptotically optimal path planning,'' \emph{IEEE Transactions on Robotics},
  vol.~34, no.~4, pp. 966--984, 2018.

\bibitem{karaman2011}
S.~Karaman and E.~Frazzoli, ``Sampling-based algorithms for optimal motion
  planning,'' \emph{The International Journal of Robotics Research (IJRR)},
  vol.~30, no.~7, pp. 846--894, 2011.

\bibitem{sakcak2019a}
B.~Sakcak, L.~Bascetta, G.~Ferretti, and M.~Prandini, ``An admissible heuristic
  to improve convergence in kinodynamic planners using motion primitives,''
  \emph{IEEE Control Systems Letters}, vol.~4, no.~1, pp. 175--180, 2020.

\bibitem{sakcak2019b}
B.~Sakcak, L.~Bascetta, G.~Ferretti, and M.~Prandini, ``Sampling-based optimal
  kinodynamic planning with motion primitives,'' \emph{Autonomous Robots},
  vol.~43, no.~7, pp. 1715--1732, 2019.

\bibitem{bellman1957}
R.~Bellman, \emph{Dynamic Programming}.\hskip 1em plus 0.5em minus 0.4em\relax
  Princeton University Press, 1957.

\bibitem{gammell2015}
J.~D. Gammell, S.~S. Srinivasa, and T.~D. Barfoot, ``{B}atch {I}nformed {T}rees
  ({BIT}*): Sampling-based optimal planning via the heuristically guided search
  of implicit random geometric graphs,'' in \emph{Proceedings of the IEEE
  International Conference of Robotics and Automation (ICRA)}.\hskip 1em plus
  0.5em minus 0.4em\relax IEEE, 2015, pp. 3067--3074.

\bibitem{gammell2020}
J.~D. Gammell, T.~D. Barfoot, and S.~S. Srinivasa, ``{B}atch {I}nformed {T}rees
  ({BIT}*): Informed asymptotically optimal anytime search,'' \emph{The
  International Journal of Robotics Research (IJRR)}, 2020.

\bibitem{penrose2003}
M.~Penrose, \emph{Random Geometric Graphs}.\hskip 1em plus 0.5em minus
  0.4em\relax Oxford University Press, 2003, vol.~5.

\bibitem{hauser2015}
K.~Hauser, ``Lazy collision checking in asymptotically-optimal motion
  planning,'' in \emph{Proceedings of the IEEE International Conference on
  Robotics and Automation (ICRA)}, 2015, pp. 2951--2957.

\bibitem{kim2018}
D.~Kim, Y.~Kwon, and S.-e. Yoon, ``Adaptive lazy collision checking for optimal
  sampling-based motion planning,'' in \emph{2018 15th International Conference
  on Ubiquitous Robots (UR)}, 2018, pp. 320--327.

\bibitem{dellin2016}
C.~M. Dellin and S.~S. Srinivasa, ``A unifying formalism for shortest path
  problems with expensive edge evaluations via lazy best-first search over
  paths with edge selectors,'' in \emph{Proceedings of the AAAI International
  Conference on Automated Planning and Scheduling (ICAPS)}, 2016.

\bibitem{mandalika2018}
A.~Mandalika, O.~Salzman, and S.~S. Srinivasa, ``Lazy receding horizon {A}* for
  efficient path planning in graphs with expensive-to-evaluate edges,'' in
  \emph{Proceedings of the AAAI International Conference on Automated Planning
  and Scheduling (ICAPS)}, 2018.

\bibitem{koenig2004}
S.~Koenig, M.~Likhachev, and D.~Furcy, ``{L}ifelong {P}lanning {A*},''
  \emph{Artificial Intelligence}, vol. 155, no. 1-2, pp. 93--146, 2004.

\bibitem{gammell2017}
J.~D. Gammell, ``Informed anytime search for continuous planning problems,''
  Ph.D. dissertation, University of Toronto, 2017.

\bibitem{janson2015}
L.~Janson, E.~Schmerling, A.~Clark, and M.~Pavone, ``{Fast} {Marching} {Tree}:
  A fast marching sampling-based method for optimal motion planning in many
  dimensions,'' \emph{The International Journal of Robotics Research}, vol.~34,
  no.~7, pp. 883--921, 2015.

\bibitem{janson2018}
L.~Janson, B.~Ichter, and M.~Pavone, ``Deterministic sampling-based motion
  planning: Optimality, complexity, and performance,'' \emph{The International
  Journal of Robotics Research}, vol.~37, no.~1, pp. 46--61, 2018.

\bibitem{aine2016}
S.~Aine and M.~Likhachev, ``Truncated incremental search,'' \emph{Artificial
  Intelligence}, vol. 234, pp. 49--77, 2016.

\bibitem{likhachev2005}
M.~Likhachev and S.~Koenig, ``A generalized framework for lifelong planning a*
  search,'' in \emph{Proceedings of the International Conference on Automated
  Planning and Scheduling (ICAPS)}, 2005.

\bibitem{likhachev2008}
M.~Likhachev, D.~Ferguson, G.~Gordon, A.~Stentz, and S.~Thrun, ``Anytime search
  in dynamic graphs,'' \emph{Artificial Intelligence}, vol. 172, no.~14, pp.
  1613--1643, 2008.

\bibitem{sucan2012}
I.~A. Sucan, M.~Moll, and L.~E. Kavraki, ``{T}he {O}pen {M}otion {P}lanning
  {L}ibrary,'' \emph{IEEE Robotics Automation Magazine}, vol.~19, no.~4, pp.
  72--82, 2012.

\bibitem{arslan2013}
O.~Arslan and P.~Triotras, ``Use of relaxation methods in sampling-based
  algorithms for optimal motion planning,'' in \emph{Proceedings of the IEEE
  International Conference on Robotics and Automation (ICRA)}.\hskip 1em plus
  0.5em minus 0.4em\relax IEEE, 2013, pp. 2421--2428.

\bibitem{thayer2012}
J.~T. Thayer, J.~Benton, and M.~Helmert, ``Better parameter-free anytime search
  by minimizing time between solutions,'' in \emph{Proceedings of the Symposium
  on Combinatorial Search (SoCS)}, 2012.

\bibitem{strub2020}
M.~P. Strub and J.~D. Gammell, ``{A}dvanced {BIT}* ({ABIT}*): {Sampling}-based
  planning with advanced graph-search techniques,'' in \emph{Proceedings of the
  IEEE International Conference on Robotics and Automation (ICRA)}, 2020.

\end{thebibliography}

\end{document}